\documentclass{article}

\usepackage{arxiv}
\usepackage[utf8]{inputenc} 
\usepackage[T1]{fontenc}    
\usepackage{url}            
\usepackage{booktabs}       
\usepackage{amsfonts}       
\usepackage{nicefrac}       
\usepackage{microtype}      
\usepackage{lipsum}
\usepackage{cite}
\usepackage{amsmath,amssymb,amsfonts}
\usepackage{graphicx}
\usepackage{textcomp}
\usepackage{comment}
\usepackage{tabularx}
\usepackage{pifont}
\newcommand{\xmark}{\ding{53}}%
\usepackage{caption}
\usepackage{subcaption}
\usepackage{array, multirow}
\usepackage{algorithm}
\usepackage{algpseudocode}
\usepackage{bm}

\usepackage{subcaption}
\usepackage{graphicx}
\graphicspath{ {./images/} }
\usepackage{graphicx}
\usepackage{amsmath}
\usepackage{amssymb}
\usepackage{booktabs}
\usepackage{multirow}
\usepackage{algpseudocode}

\usepackage[pagebackref,breaklinks,colorlinks]{hyperref}
\usepackage[capitalize]{cleveref}

\title{Augmented Efficiency: Reducing Memory Footprint and Accelerating Inference for 3D Semantic Segmentation through Hybrid Vision}

\author{
 Aditya Krishnan$^1$, Jayneel Vora$^1$, Prasant Mohapatra$^2$\\
$^{1}$Department of Computer Science, University of California, Davis\\
$^{2}$Department of Computer Science, University of South Florida, Tampa\\
\texttt{\{adikrishnan,jrvora\}@ucdavis.edu, pmohapatra@usf.edu}
\thanks{$^1$ Equal contribution.}
}

\begin{document}
\maketitle

\begin{abstract}
Semantic segmentation has emerged as a pivotal area of study in computer vision, offering profound implications for scene understanding and elevating human-machine interactions across various domains. While 2D semantic segmentation has witnessed significant strides in the form of lightweight, high-precision models, transitioning to 3D semantic segmentation poses distinct challenges. Our research focuses on achieving efficiency and lightweight design for 3D semantic segmentation models, similar to those achieved for 2D models. Such a design impacts applications of 3D semantic segmentation where memory and latency are of concern. This paper introduces a novel approach to 3D semantic segmentation, distinguished by incorporating a hybrid blend of 2D and 3D computer vision techniques, enabling a streamlined, efficient process.

We conduct 2D semantic segmentation on RGB images linked to 3D point clouds and extend the results to 3D using an extrusion technique for specific class labels, reducing the point cloud subspace. We perform rigorous evaluations with the DeepViewAgg model on the complete point cloud as our baseline by measuring the  Intersection over Union (IoU) accuracy, inference time latency, and memory consumption. This model serves as the current state-of-the-art 3D semantic segmentation model on the KITTI-360 dataset. We can achieve heightened accuracy outcomes, surpassing the baseline for 6 out of the 15 classes while maintaining a marginal 1\% deviation below the baseline for the remaining class labels. Our segmentation approach demonstrates a 1.347x speedup and about a 43\% reduced memory usage compared to the baseline. 
\end{abstract}


\section{Introduction}
\label{sec:introduction}
\begin{figure}[t]
  \centering
     \begin{subfigure}[b]{0.45\columnwidth}
         \centering
         \includegraphics[width=\columnwidth]{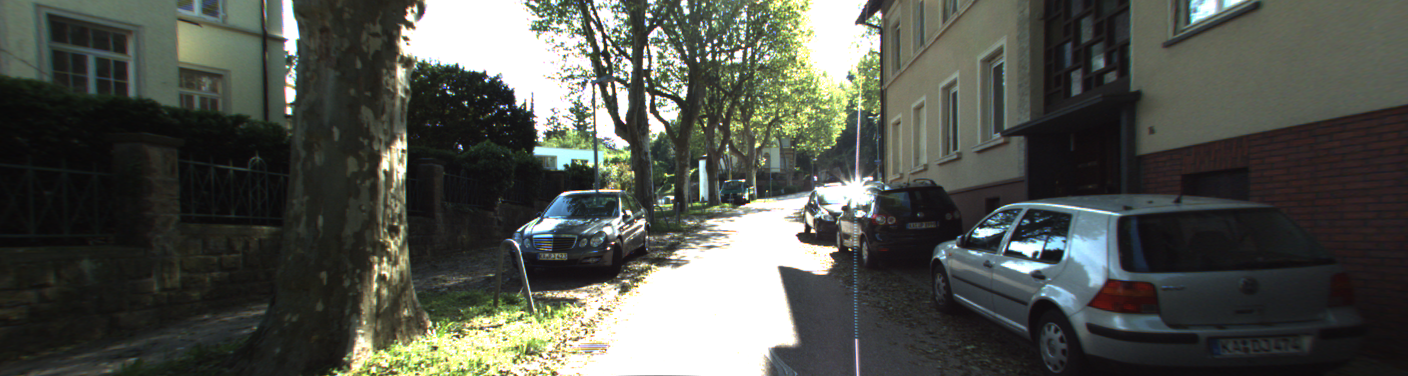}
         \caption{2D Raw Data}
         \label{fig:2draw}
     \end{subfigure}
     \begin{subfigure}[b]{0.45\columnwidth}
         \centering
         \includegraphics[width=\columnwidth]{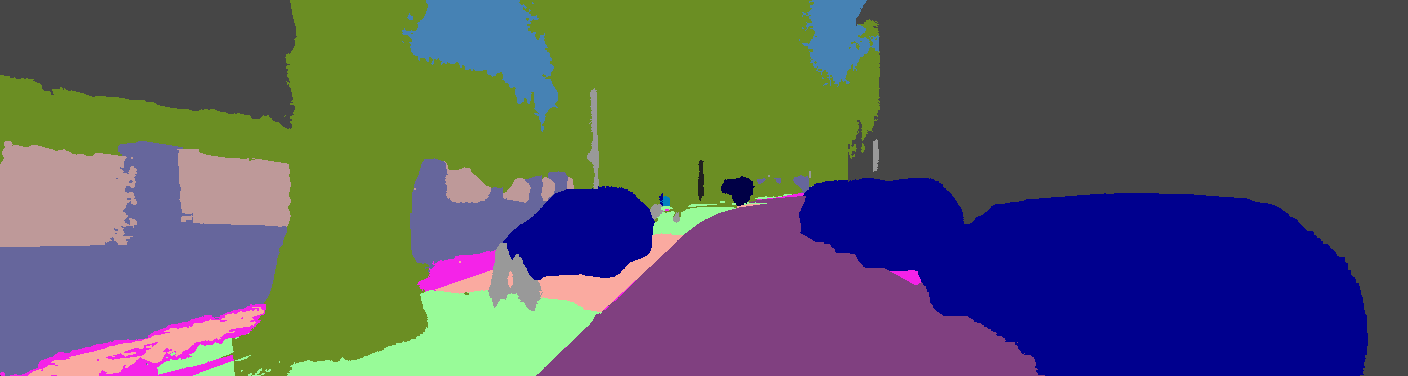}
         \caption{2D Semantic Ground-Truth}
         \label{fig:2dsem}
     \end{subfigure}
     \vfill
     \begin{subfigure}[b]{0.45\columnwidth}
         \centering
         \includegraphics[width=\columnwidth]{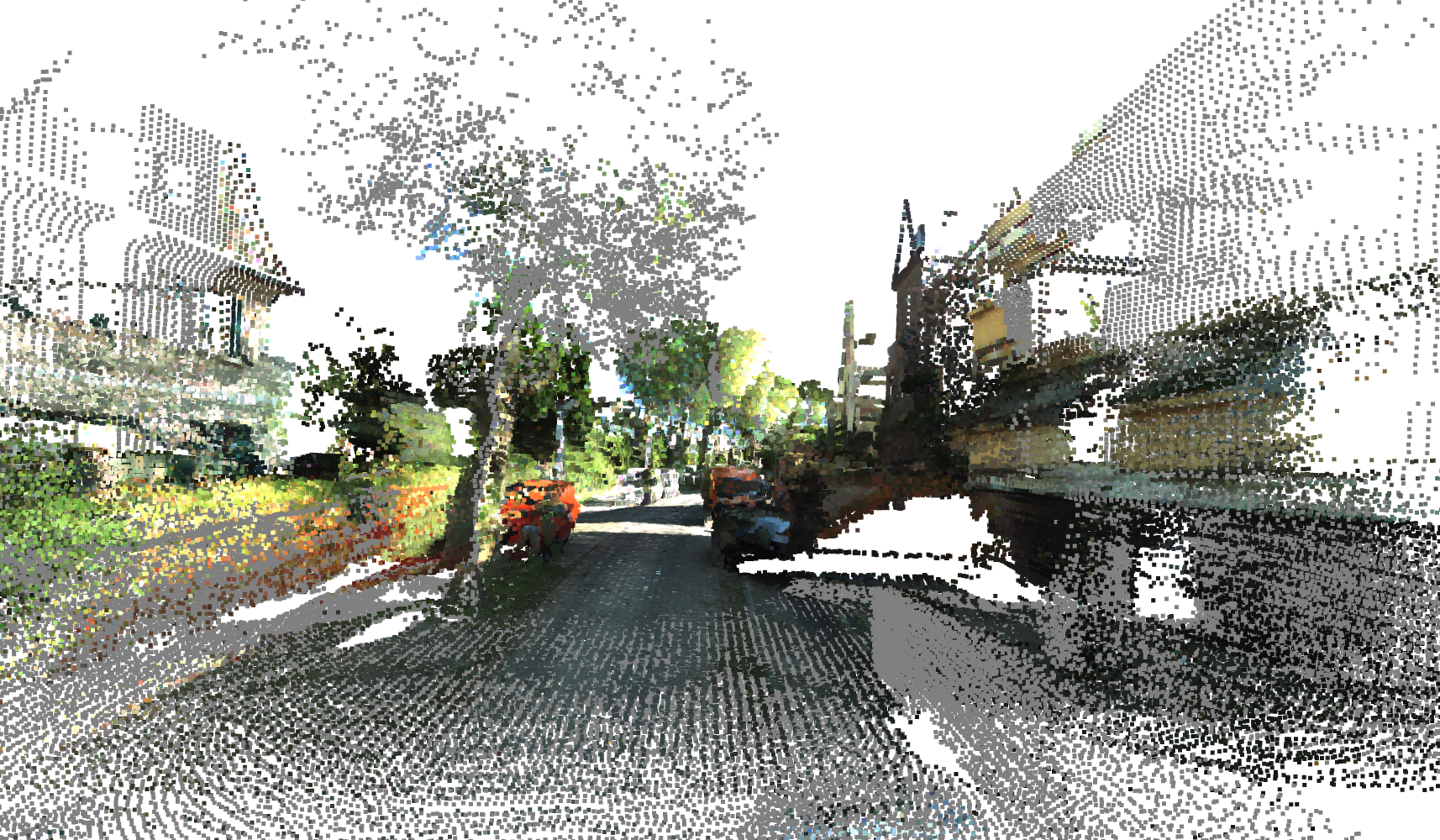}
         \caption{3D Raw Data}
         \label{fig:3draw}
     \end{subfigure}
     \begin{subfigure}[b]{0.45\columnwidth}
         \centering
         \includegraphics[width=\columnwidth]{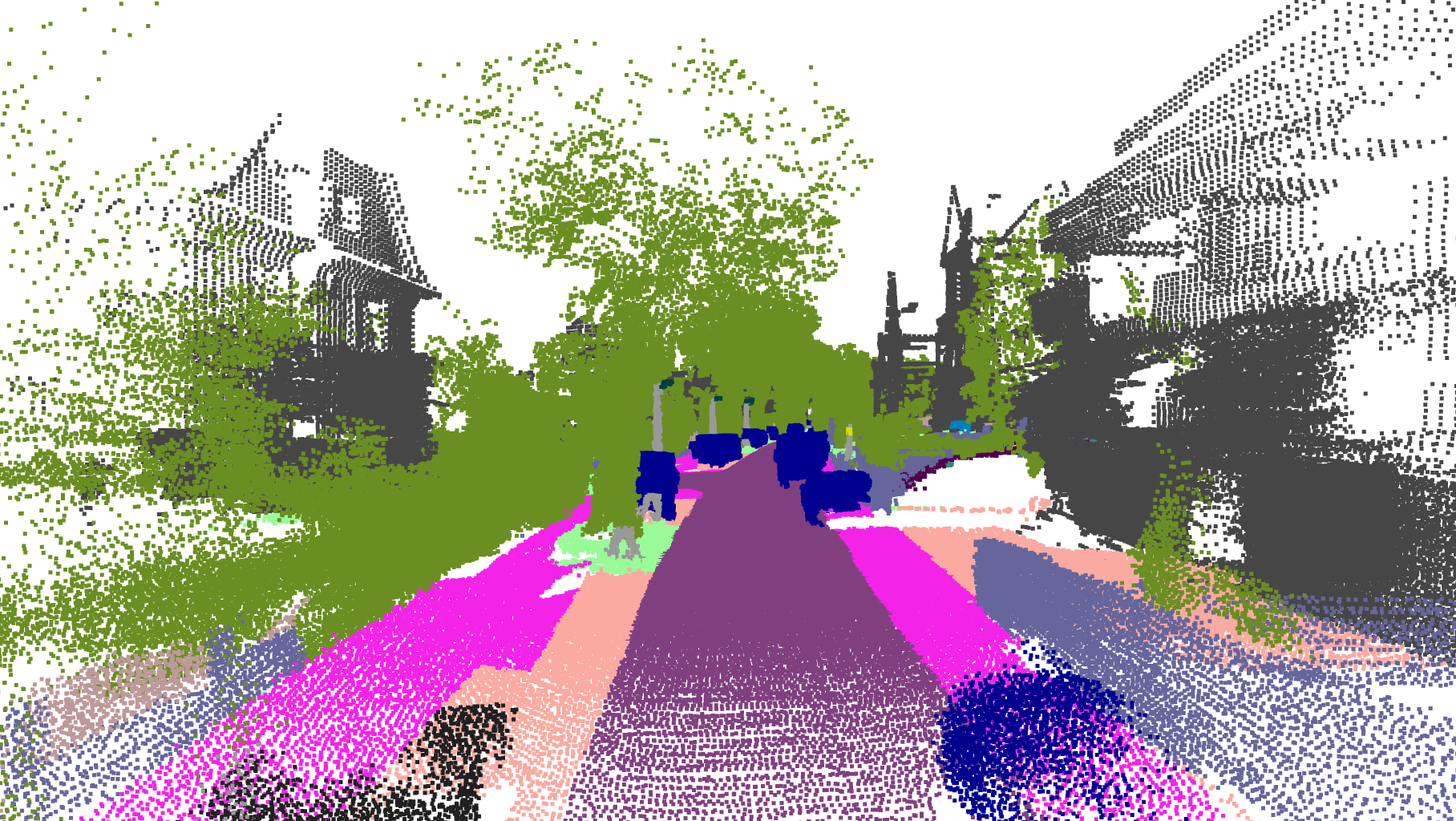}
         \caption{3D Semantic Ground Truth}
         \label{fig:3dsem}
     \end{subfigure}
         \caption{KITTI-360 raw training data and corresponding ground truth}
    \label{fig:semseg}
\end{figure} 
Research in computer vision has established a widespread need to understand and process the information around us meaningfully through computational capabilities. Semantic segmentation is an example of such a mechanism for meaningfully understanding this information in a manner that promotes visual acuity. Semantic segmentation is a pixel-level classification task where an image is segmented into semantically meaningful partitions. The partitions reflect spatial relationships among objects and identify object boundaries.

Semantic segmentation is pivotal in computer vision with various applications across diverse domains. 
In the domain of autonomous vehicles, semantic segmentation is indispensable for real-time scene understanding and object detection, facilitating safer navigation and collision avoidance \cite{milioto2019range, yang2021skim, scene_app, superresolution_app, hmi_app}. In agriculture, semantic segmentation assists in crop monitoring and yield estimation, enabling farmers to optimize resource utilization and improve productivity \cite{christensen2017semantic, wu2017deep}.
 
Within medical imaging, semantic segmentation aids in the precise delineation of anatomical structures and pathological regions, empowering doctors with critical insights for accurate diagnosis, prognosis, and treatment planning \cite{pratt2006region, perez2018deep}. 
Semantic segmentation is a fundamental component in robotics and robotic vision, allowing robots to perceive their surroundings and make informed decisions for tasks like object manipulation and navigation \cite{fang2020autonomous, liu2021distributed}. The applications of semantic segmentation are equally significant in augmented reality, enhancing the user experience by seamlessly integrating virtual and real-world elements \cite{mei2019semantic, jiang2021interactive}. 
These use cases underscore the versatility and importance of semantic segmentation, empowering various industries with advanced computer vision capabilities.

Point clouds are a comprehensive data representation of 3D scenes, comprising individual points with associated coordinates, color, and intensity information. 
This representation offers a comprehensive representation of the original geometric structure \cite{electronics12173642}. 
Figure \ref{fig:semseg} depicts snapshots of the raw and ground truth data for both 2D and 3D data from the KITTI-360 dataset.
This figure showcases 2D images from the training dataset provided by KITTI-360 and 3D training data images obtained using visualization scripts provided by KITTI-360 in \cite{kitti360}.
In transitioning from 2D to 3D semantic segmentation, we navigate challenges like partial occlusion, depth perception, lighting variations, and intricate 3D object structures. While the richer representation enhances accuracy, it escalates computational demands, especially concerning 3D volumetric data analysis. This issue gains urgency in mobile device contexts, emphasizing the demand for innovative 3D semantic segmentation solutions given constraints like low power, limited resources, and real-time processing.
Sparsifying data in 3D semantic segmentation accelerates processing and boosts accuracy by focusing computational resources on relevant points within volumetric datasets. This valuable asset advances 3D semantic segmentation models in computer science, AI research, and applications.

This work proposes a valuable 2D to 3D extrusion process to be applied atop the DeepViewAgg model flow, a 3D semantic segmentation model. Our contributions encompass the following:

\begin{itemize}
    \item We propose a novel framework that enhances the computational and memory efficiency of 3D semantic segmentation.
    \item Our approach leverages 2D semantic segmentation as an initial step, strategically reducing the density of points within the point cloud required for subsequent 3D semantic segmentation.
    \item We assess the effectiveness of our strategy by comparing it to DeepViewAgg, our baseline model, on the KITTI-360 dataset. The results demonstrate superior accuracy for specific class labels, with a marginal deviation from baseline accuracy for others.
    \item Furthermore, we investigate the potential of our hybrid 2D/3D vision approach to improve the efficiency of 3D semantic segmentation, while maintaining accuracy comparable to the baseline.
\end{itemize}

Our approach achieves superior IoU accuracy results, surpassing our baseline for 6 out of the 15 class labels [2, 6, 10, 11, 12, 14]. For the remaining nine classes, we maintain IoU accuracy within a marginal 1\% deviation below the baseline. In addition, our approach achieves a 1.347x speed up for our segmentation process and at least a 43\% reduction in memory usage compared to the baseline.

The following sections outline the organization of this paper: Section \hyperref[sec:RWM]{2} explores related work and outlines the motivation behind our proposal. 
Section \hyperref[sec:M]{3} discusses in detail the algorithms and corresponding modules we propose in this paper to perform accelerated resource-efficient 3D semantic segmentation and the semantic segmentation models used in this process. 
Section \hyperref[sec:E]{4} reviews the dataset upon which we run our tests, the metrics we use to evaluate our methodology, the baseline against which we compare our performance, and the device information upon which we ran our evaluations. 
Section \hyperref[sec:RD]{5} describes the results obtained from our process and the baseline and discusses these results for each metric we use to compare our approach and the baseline. 
Section \hyperref[sec:CFW]{6} contains a conclusion and an overview of future work.

\section{Related Work \& Motivation}
\label{sec:RWM}
\subsection{Semantic Segmentation}
\subsubsection{2D Semantic Segmentation}
Many works have aimed to make  2D semantic segmentation models run efficiently. 
The common methods attempt to improve the inference speeds of their model through two main approaches: lightweight backbones and shorter network structures.
As examples of the first type, JPANet \cite{JPANet} and DFANet \cite{DFANet} adopt GhostNet \cite{Ghostnet} and Xception \cite{xcept} as backbone network extraction features, respectively. This approach allows for high semantic segmentation accuracy with low inference speeds; however, the application scopes for these models are limited due to increased computational complexity.

ENet \cite{ENet}, ESPNet \cite{ESPNet}, and FastSCNN \cite{SCNN} achieve high inference speeds using short network structures but suffer from decreased segmentation accuracy.
More recently, there has been interest in developing 2D semantic segmentation models explicitly built for mobile device frameworks.
For example, LMANet \cite{LMANet}, a multi-scale attention-guided network based on asymmetric encoder-decoder, and RAFNet \cite{RAFNet}, a reparameterizable across-resolution fusion network, have both been recently published as models that aim to perform real-time segmentation while improving inference speed and achieving high segmentation accuracy.
In addition, MobileNetV2 \cite{mobnetv2}, and later, MobileNetV3 \cite{mobnetv3}, establish themselves as neural network models built specifically for mobile device deployment with the ability to perform object detection, object classification, and semantic segmentation in 2D while maintaining lightweight resource-efficient performance.

\subsubsection{3D Semantic Segmentation}
Multiple strategies have been explored to improve efficiency and resource consumption of 2D semantic segmentation for various applications and hardware environments. However, extending these efforts to develop lightweight and resource-efficient 3D semantic segmentation 
has yet to be adequately addressed. Previous research has examined feature networks for pointwise semantic segmentation of 3D urban scenes from mobile terrestrial and airborne LiDAR point clouds \cite{isprs-2019}. The Slimmer approach \cite{slimmer} aims to accelerate 3D semantic segmentation for mobile augmented reality by reducing the points in the point cloud under a K-Nearest Neighbor model. Another study by Cadena et al. \cite{hetersensor} employs a Conditional Random Field to segment point clouds from an RGB-D sensor, merging segmentation results to enhance accuracy while minimizing computational costs. PointSeg \cite{pointseg}, a real-time semantic segmentation method for road objects based on 3D LiDAR point clouds, converts data into spherical input for an updated SqueezeSeg model. However, these methods exhibit limitations, such as specialized application scopes, high resource consumption to improve accuracy, loss of accuracy in exchange for improved resource efficiency, or a restricted focus on class studies.
DeepViewAgg \cite{deepviewagg} an approach for 3D semantic segmentation that is dependent on 2D images, utilizes multiple 2D views of the data and maps these image pixels to the points in the 3D point clouds based on visibility. This approach involves aggregating information from multiple 2D views to perform 3D semantic segmentation, leveraging the strengths of 2D semantic models for 3D scene understanding. However, DeepViewAgg is dependent on increased processing associated with aggregating multiple 2D views, is dependent on the availability of large amounts of 2D data, and is limited in applications that require real-time segmentation. We mitigate this limitation by performing 2D semantic segmentation on the 2D views allowing us to focus the 3D semantic segmentation classification process on the 3D point cloud without the need for large amounts of 2D images with great quality.

\subsection{Hybrid Vision}
The research on improving 3D semantic segmentation within mobile frameworks needs improvement. Recent proposals for hybrid vision techniques in mobile frameworks aim to combine the speed of 3D vision with the resource efficiency of 2D vision for 3D computer vision tasks. However, existing research primarily concentrates on hybrid vision for 3D object detection and classification \cite{hybridvis}. Approaches like DeepMix \cite{deepmix} and RGB-D image data analysis using MATLAB \cite{rgbd} target mobile AR headsets, emphasizing object detection. Liu et al. \cite{hybridvis} have recently theorized a hybrid vision approach for semantic segmentation, but this area remains largely unexplored.

\subsection{Motivation}
Semantic segmentation in 3D is becoming increasingly important and useful in contexts where 2D semantic segmentation is incapable of providing adequate information for scene understanding.
Despite the growing interest in 3D semantic segmentation, there remains a significant gap in research dedicated to creating lightweight and efficient applications. 
Current methodologies aim to impact high-performance computing systems, leveraging substantial computational and memory resources. 
Existing work is limited in its ability to reconcile the demands of accuracy and resource efficiency for 3D semantic segmentation.
This gap underscores the necessity for innovative solutions that reconcile the demands of accuracy and resource efficiency in 3D semantic segmentation.

Furthermore, while hybrid vision techniques have marked significant advancements in augmented reality (AR) technologies and 3D object detection/classification, their potential in 3D semantic segmentation remains under-explored. 
The majority of existing works, such as those discussed in \cite{hybridvis}, have concentrated on the theoretical underpinnings or applications in contexts that do not prioritize computational efficiency or low-memory usage, especially for 3D semantic segmentation.

The state-of-the-art approach for 3D semantic segmentation is dependent on performing segmentation on the complete point cloud regardless of the needs of the user.
In many cases, users of 3D semantic segmentation algorithms only require segmentation of select classes as they are important to the specific scene understanding goals of the user.
We aim to mitigate this challenge by providing this control to the user, identifying the select classes the user aims to segment from the point clouds and focusing our processing on the data that are classified as such.

This paper sets out to bridge these gaps by introducing a novel method that leverages hybrid vision techniques for 3D semantic segmentation. 
Our approach is distinguished by its focus on computational and memory efficiency, aiming to significantly reduce the resource footprint of point cloud semantic segmentation without compromising, and in some cases improving, the accuracy of segmentation results. 
This is achieved through a careful integration of 2D and 3D semantic segmentation algorithms with the goal of improving the data handling of 3D point cloud data for semantic segmentation without the need to handle the complete point cloud.
Through this work, we contribute to the emerging field of hybrid vision for 3D semantic segmentation by providing a practical framework that addresses the critical limitations of prior work.
Ultimately we aim to pave the way for future research and applications that are both efficient and effective.

\section{Method}
\label{sec:M}
\subsection{System Model: Inputs}
In 2D semantic segmentation, the input data typically consists of 2D images represented as 2D arrays, where each element corresponds to a pixel in the 2D image.
These images can take various forms, such as RGB, grayscale, or multispectral images. 
In the case of RGB images, each pixel is represented by three color channels (red, green, and blue), whereas grayscale images use only one channel, representing pixel intensity.
A few key attributes of the input data for 2D semantic segmentation include the following:

\begin{itemize}
    \item \textbf{Image Resolution}: The spatial resolution of the image, generally measured in pixels or pixels per unit length.
    \item \textbf{Number of Channels}: For multi-channel images, the channels or bands contain different information types depending on the pixels' format.
    \item \textbf{Pixel Values}: The values of individual pixels could range from 0 to 255 for 8-bit images or be normalized to a different scale depending on the use-case of the 2D semantic segmentation model in question.
\end{itemize}

In 3D semantic segmentation, the input data consists of point clouds or voxelized representations of 3D scenes. 
A voxel, otherwise known as a volumetric pixel, is simply a pixel that contains color information depending on the format of the image pixels and the position related to spatial localization.
The input data for 3D semantic segmentation can be represented in two primary ways:

1. \textbf{Point Clouds}: Point clouds are collections of 3D points in space, where each point is described by its (x, y, z) coordinates. 
LiDAR (Light Detection And Ranging) or structured light scanning technologies can capture these point clouds.

2. \textbf{Voxelized Data}: Voxelized data represents the 3D scene using a grid structure of cubic elements called voxels.
In some implementations for 3D semantic segmentation, these voxels do not necessarily need to be cubic but can take on any other type of volumetric geometry, like spherical or cylindrical.
Each voxel can store information about the objects they contain or their properties within a volume element.

A few key attributes of the input data for 3D semantic segmentation include:

\begin{itemize}
    \item \textbf{Number of Points}: In point clouds, the number of 3D points can vary widely based on the sensors used and the scene complexity.
    \item \textbf{Point Features}: Additional attributes accompanying with each point in the point cloud, such as color, intensity, or reflectance. 
    These attributes are present depending on the sensors used to obtain the data in the point cloud.
    \item \textbf{Voxelization Parameters}: For voxelized data, parameters such as the voxel size and grid resolution can affect the granularity of the representation.
    Voxel size and grid resolution can affect the semantic segmentation accuracy but are ultimately determined by the model requirements and other factors related to the performance of the segmentation task.
    In general, the application for which segmentation is performed can also dictate the resolution of these parameters.
\end{itemize}

\subsection{System Model: Model Structure}
The structure of a 2D semantic segmentation model is designed to process 2D image data and produce class predictions for each pixel in the image. Usually, this process of segmentation is broken down as follows:

\begin{itemize}
 \item \textbf{Input Structure:} The input image $I$ is a 2D array represented as a matrix. Each pixel is notated by its position $(x, y)$ and its color or intensity values.
 \item \textbf{Convolutional Layers:} The model often includes convolutional layers to extract hierarchical features. Let $fmap_{2D}(I)$ represent the feature map generated by the convolutional layers \cite{ronneberger2015semantic, chen2018deeplab}.
 \item \textbf{Mapping Transform:} A mapping transform, often implemented through a decoder, refines features and produces the final pixel-wise class predictions. Let $P_{2D}(I)$ represent the predicted pixel-wise class probabilities \cite{ronneberger2015u}.
 \item \textbf{Loss Function:} The loss function for 2D semantic segmentation can be represented as:

\[
\mathcal{L}_{2D} = \sum_{x,y} \text{CrossEntropy}(P_{2D}(I), \text{GroundTruth}(I, x, y))
\]

where, $\text{GroundTruth}(I, x, y)$ is the ground truth label for pixel $(x, y)$ \cite{ronneberger2015semantic}.
This data is provided by the dataset used for training and evaluating the model \cite{ronneberger2015semantic}.
\end{itemize}

3D semantic segmentation models are designed to process 3D point clouds or voxelized data to produce volumetric class predictions. Generally, this process involves the following:

\begin{itemize}
    \item \textbf{Input Structure:} The input data for 3D semantic segmentation includes point clouds or voxelized 3D grids, represented as a collection of points or voxels in 3D space. In some cases, the raw data from the point clouds is processed using linear transforms before performing voxelization to the processed point cloud data to be then inputted to the neural network model \cite{qi2017pointnet, maturana2015voxnet}.
    \item \textbf{Convolutional Layers:} 3D semantic segmentation models utilize specialized 3D convolution layers to capture spatial information. Let $fmap_{3D}(S)$ represent the 3D feature map generated by these layers \cite{qi2017pointnet, maturana2015voxnet}.
    \item \textbf{Mapping Transform:} The model produces volumetric class predictions, often represented as a 3D semantic label volume. Let $V_{3D}(S)$ represent this predicted volume \cite{qi2017pointnet, maturana2015voxnet}.
    \item \textbf{Loss Function:} The loss function for 3D semantic segmentation can be represented as:

\[
\mathcal{L}_{3D} = \sum_{i,j,k} \text{CrossEntropy}(V_{3D}(S), \text{GroundTruth}(S, i, j, k))
\]

Here, $\text{GroundTruth}(S, i, j, k)$ is the ground truth label for voxel $(i, j, k)$.
Again, this data is provided by the dataset used for training and evaluating the model \cite{qi2017pointnet, maturana2015voxnet}.
\end{itemize}

\subsection{Semantic Segmentation: 2D vs. 3D}
Semantic segmentation in 2D and 3D differ significantly in terms of dimensionality and representation, leading to unique challenges and applications, as follows:

- \textit{Dimensionality}: 2D semantic segmentation operates on 2D images, while 3D semantic segmentation processes 3D point clouds or voxelized data. 
The dimension difference promotes an expansion in the quantity of data to process, which can drastically affect the model's runtime and resources consumed to process and perform segmentation on the data.

- \textit{Spatial Relationships}: 3D semantic segmentation captures spatial relationships in the third dimension, essential for understanding volumetric data.
This increased understanding establishes a means for a better understanding of the occlusion of objects, the relative distances between objects, and context-based relationships between objects that are difficult to perceive through 2D semantic segmentation.

- \textit{Data Volume}: 3D data often involves larger data volumes, presenting computational complexity and memory usage challenges.
These challenges are what we aim to tackle and mitigate in the approach proposed by this paper. 

- \textit{Applications}: 2D semantic segmentation is widely used in image analysis, while 3D semantic segmentation is proper for applications in diverse fields, including autonomous vehicles, robotics, and 3D scene understanding. 3D semantic segmentation also offers several advantages and improvements over its 2D counterpart. These improvements are particularly significant in applications where understanding the three-dimensional structure of the scene is crucial:

- \textit{Spatial Awareness}: 3D semantic segmentation captures scenes' volumetric nature, allowing for improved spatial awareness. 
Spatial awareness is essential in applications like autonomous driving, where understanding objects' height, depth, and orientation is crucial, and in healthcare, for example, endoscopic tumor image analysis.

- \textit{Enhanced Realism}: 3D semantic segmentation contributes to more realistic 3D reconstructions and virtual environments, as it accurately models the depth and layout of objects.
Applications such as augmented and mixed reality benefit from this enhanced understanding of scenes, allowing for improved human-machine interactions.

- \textit{Precise Object Localization}: 3D semantic segmentation enables more precise object localization in 3D space, vital in robotics and augmented reality.
Such applications utilize this localization to improve navigation and more fine-tuned machine-scene interaction using spatial relationships between objects.

- \textit{Better Scene Understanding}: In complex 3D scenes, 3D semantic segmentation outperforms 2D semantic segmentation by providing richer information about object occlusions and scene geometry.
This improved scene understanding allows for innovative solutions to scene-related problems while establishing a mechanism for fine-tuned scoping of applications for segmentation, especially in scenarios where physical interaction with the scene is limited for humans.

- \textit{Expanded Applications}: 3D semantic segmentation extends the range of interdisciplinary applications, including geospatial analysis, archaeology, medical imaging, and more, where 2D semantic segmentation is insufficient.

3D semantic segmentation has provided significant contributions in various domains where understanding the three-dimensional aspects of data is paramount.
However, as discussed above, the utility of 3D semantic segmentation is limited in the realm of low-power, low-resource devices that would benefit from performing these tasks quickly.

\subsection{3D Semantic Segmentation Process} 

\begin{figure}[htp]
\centering
\includegraphics[width=\textwidth]{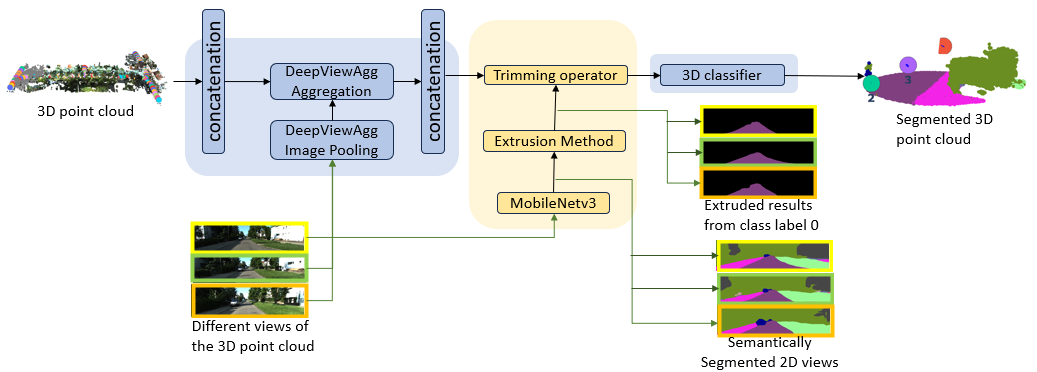}
\caption{Schematic Representation of Our Novel Integration Process. This diagram illustrates the unique fusion of multi-view 2D semantic segmentation and 3D point cloud data refinement, showcasing the extrusion for the road class as an example.}
\label{fig:process}
\end{figure}

This work introduces a pioneering approach to 3D semantic segmentation, encapsulated in Figure \ref{fig:process}, which extends beyond traditional methodologies by innovatively combining 2D semantic segmentation insights with 3D point cloud analysis. Existing work in semantic segmentation has not considered implementing a solution through hybrid vision which our approach aims to address. Our process, inspired by and significantly advancing the DeepViewAgg model's framework, introduces a novel Aggregation Block that synergizes multi-perspective data, enhancing the depth and reliability of segmentation outcomes.

Distinguishing itself from mere data preprocessing, our methodology employs a sophisticated Image Pooling Block, adapted from DeepViewAgg, for efficient batch processing of multi-view imagery directly linked to corresponding point clouds. The image views are associated with points in the corresponding point cloud data by assessing the visibility of points by pixels in the image views considered. The critical innovation lies in our bespoke 2D/3D extrusion algorithm, which meticulously extracts and elevates 2D semantic segmentation results, performed by the state-of-the-art MobileNetV3, into the 3D domain. By utilizing 2D semantic segmentation on image views and extracting visible pixel-point pairs, we can mitigate the limitation of abundance of data associated with the traditional DeepViewAgg process flow. Specifically, we obtain all points in the segmented 2D image view to obtain points necessary to keep in extrusion to the 3D point cloud. This way only points in the point cloud that are mapped to pixels in the image views that correspond to the necessary class label are kept when extruding the pixels into 3D space. In addition, we ensure applicability of our approach is directly dependent on the 2D semantic segmentation output which addresses the concerns of image quality when considering the availability and quality of image views. This strategic extrusion, as outlined in Algorithm \ref{algo:2d3dextrusion}, focuses the segmentation process on areas of the point cloud most likely to embody the designated class labels. This process refines the accuracy and efficiency of subsequent 3D semantic segmentation.

\begin{algorithm}
\caption{2D/3D Extrusion Algorithm}
\label{algo:2d3dextrusion}
\textbf{Input:} 
\begin{itemize}
    \item $V$: List of 2D images for validation.
    \item $T$: List of 2D images for the test dataset.
    \item HEIGHT: Height of images in dataset measured in pixels.
    \item WIDTH: Width of images in dataset measured in pixels.
\end{itemize}

\textbf{Output:} 
\begin{itemize}
    \item JSON files containing dictionaries with class labels as keys and lists of pixels as values.
\end{itemize}

\begin{algorithmic}
\For{each image $i$ in $V$ or $T$}
    \For{each class label $cl$}
        \State Extract a list of pixels $[px, py]$ associated with class label $cl$ in image $i$, where  $px \in [0,HEIGHT-1]$, and $py \in [0,WIDTH-1]$.
        \State Store a dictionary with the class label as the key and list of pixels as the value and store into a JSON file named similar to the image's file name.
    \EndFor
\EndFor
\end{algorithmic}
\end{algorithm}

Furthermore, our 3D Semantic Segmentation Classifier Block represents a significant leap forward, not only by refining segmentation on precisely trimmed point cloud data but by redefining the workflow of integrating 2D insights into 3D semantic segmentation landscapes. By integrating the 2D extruded outputs into the 3D point cloud data processed by the 3D Classifier Block, we ensure streamlined classification focused on data associated with the class label(s) deemed important for classification by the user. This ensures more efficient 3D semantic segmentation processing with reduced memory overhead, limited accuracy loss, and, in some cases, reasonable accuracy gain. Algorithm \ref{algo:preprocess} outlines this process by updating the point-image pairs obtained in the aggregation and pooling blocks of the DeepViewAgg process flow to include the data corresponding only to pixels extruded from the images by Algorithm \ref{algo:2d3dextrusion}.

\begin{algorithm}
\caption{: 3D Extrusion Preprocessing}
\label{algo:preprocess}
\textbf{Input:} 
\begin{itemize}
    \item $V$: List of 2D images for validation.
    \item $T$: List of 2D images for the test dataset.
    \item $data$: Dictionary of all data corresponding to the point cloud as provided in the dataset. $data.pos$ contains the points in the point cloud formatted as a point $(x, y, z)$.
\end{itemize}
\textbf{Output:} 
\begin{itemize}
    \item $image\_ids$: a list of indices corresponding to images in $images.paths$.
    \item $point\_ids$: a list of indices corresponding to $data.pos$.
    \item $images\_pixels$: a list of pixels with one-to-one correspondence to $image\_ids$.
\end{itemize}
\begin{algorithmic}
\State Let $images.paths$ be the list of image paths corresponding to the filepath of the image file utilizing the naming convention adopted by the dataset.
\State Let $image\_ids$ be the list of indices corresponding to images in $images.paths$.
\State Let $images\_pixels$ be a list of pixels with one-to-one correspondence to $image\_ids$. These pixels are the pixels considered by the DeepViewAgg process flow for each image view mapped to the point cloud data.
\State Let $point\_ids$ be the list of indices corresponding to $data.pos$
\State Initialize empty list $point\_data\_idx$.
\\
\State Parse the paths in $images.paths$ by folder and filename into $filepaths$. \\
\State Load all pixel lists for each filename in $filepaths$ into the $label\_dict$ by dumping the JSON file output from Algorithm \ref{algo:2d3dextrusion}.

\For{each $pix\_id$, $pixel$ in $images\_pixels$}
    \State Let $file$ be the filename obtained from $filepaths$ \\
    \hspace{2.25em} corresponding to $pix\_id$
    \If{$pixel\_list$ for $file$ is not empty \textbf{and} $pixel$ exists \\
    \hspace{2.25em} in $pixel\_list$} 
    \State Append $pix\_id$ into $point\_data\_idx$.
    \EndIf
\EndFor
\\
\State Update data in $image\_ids$, $point\_ids$, and $images\_pixels$ corresponding to the $pix\_id$s in $point\_data\_idx$. 
\end{algorithmic}
\end{algorithm}

Our approach reimagines the segmentation pipeline, ensuring that the 3D semantic segmentation algorithm operates on highly relevant and optimized datasets, as facilitated by our preprocessing innovations and the integration of 2D semantic segmentation in the 3D semantic segmentation workflow. This enables our method to achieve similar segmentation accuracy to the traditional DeepViewAgg segmentation approach, dramatically reducing computational overhead and memory consumption compared to existing methodologies.

Critically, this paper's contribution transcends conventional preprocessing by offering a comprehensive and scalable solution to 3D semantic segmentation challenges. Our integration of 2D semantic segmentation precision with 3D data analysis paves the way for more accurate, efficient, and versatile applications, setting a new standard for research in the field.

\subsection{2D Model: MobileNetV3}
The selection of the MobileNetV3 model as the 2D semantic segmentation model in our process was based on several key factors.

MobileNetV3 is renowned for its efficiency, making it a suitable choice for mobile applications and scenarios with limited computational resources. 
The model strikes a balance between high segmentation accuracy and low resource utilization. 
It achieves this by leveraging a relatively small number of parameters while maintaining a high Frames Per Second (FPS) efficiency.

MobileNetV3 builds upon the foundation of its predecessor, MobileNetV2, by introducing a new block in the network architecture. 
This block incorporates a hard-sigmoid activation function, which significantly enhances computational efficiency. 
This enhancement is particularly valuable when dealing with fixed-point arithmetic, as it simplifies the maintenance of accuracy.

MobileNetV3 maintains a network structure that aligns with its predecessor, MobileNetV2, utilizing a linear bottleneck and an inverted residual structure. 
These architectural choices efficiently address the low-rank nature of the problem, optimizing layer structures.

MobileNetV3 incorporates a variety of novel features to improve its performance. 
Notably, it introduces a new residual layer that implements a squeeze-and-excite mechanism and utilizes the hard-sigmoid activation. 
Additionally, the model employs modified swish nonlinearities based on previous works \cite{Ramachandran2018SearchingFA, ELFWING20183, Hendrycks2016BridgingNA} to enhance its layers.

MobileNetV3 employs platform-aware network architecture systems (NAS) to optimize global network structures. 
It leverages the NetAdapt algorithm for per-layer filter optimization. 
This platform-aware NAS approach is similar to that used in MnasNet-A1 \cite{mnasnet}. 
The NetAdapt algorithm fine-tunes individual layers sequentially by minimizing the latency and accuracy change ratio.

MobileNetV3 offers two variants, MobileNetV3-Large and MobileNetV3-Small, catering to high and low-resource use cases. 
We opted for MobileNetV3-Small for this work to minimize resource usage and memory overhead.

In \cite{mobnetv3}, they compared MobileNetV3 with its predecessor, MobileNetV2, as a network backbone for semantic segmentation. 
They also evaluated two segmentation heads: Reduced Atrous Spatial Pyramid Pooling (R-ASPP) and a lightweight R-ASPP (LR-ASPP). 
The LR-ASPP, inspired by the squeeze-and-excitation module in MobileNetV3, enhances performance by implementing global-average pooling.

Their experiments on the Cityscapes dataset employed mean Intersection over Union (mIOU) as the accuracy metric. 
As shown in Table \ref{tab:mobnetres}, adapted from the original MobileNetV3 paper \cite{mobnetv3}, the results indicate that MobileNetV3 exhibits competitive and comparable performance with other state-of-the-art models. 
Notable findings include the improved computation speed by reducing channels in the final block of the network backbone and the enhanced performance of the proposed LR-ASPP segmentation head.

In summary, the selection of MobileNetV3-Small as our 2D model was driven by its lightweight nature, high accuracy, and resource-efficient processing, making it an ideal choice for our 2D semantic segmentation process.
\begin{table*}[t]
\centering
  \caption{Semantic segmentation results on Cityscapes val set. \textit{RF2}: \textit{R}educe the \textit{F}ilters in the last block by a factor of \textit{2}. V2 0.5 and V2 0.35 are MobileNetV2 with depth multipliers = 0.5 and 0.35, respectively. \textit{SH}: \textit{S}egmentation \textit{H}ead, where × employs the R-ASPP while X employs LR-ASPP proposed in \cite{mobnetv3}. \textit{F}: Number of \textit{F}ilters used in the Segmentation Head. \textit{CPU (f)}: CPU time measured on a single large core of Pixel 3 (floating point) w.r.t. a \textit{full-resolution} input (i.e., 1024 × 2048). \textit{CPU (h)}: CPU time measured w.r.t. a \textit{half-resolution} input (i.e., 512 × 1024). Row 8 and 11 are the MobileNetV3 segmentation candidates. \textit{This caption and table are adapted with permission from the original MobileNetV3 paper in \cite{mobnetv3}}.}
  \label{tab:mobnetres}
  \normalsize
  \begin{tabular}{l l | c c c | c c c c c}
    \toprule
    N & Backbone & RF2 & SH & F & mIOU & Params & MAdds & CPU (f) & CPU (h)\\
    \midrule
    1 & V2 & - & \xmark & 256 & 72.84 & 2.11M & 21.29B & 3.90s & 1.02s\\
    2 & V2 & \checkmark & \xmark & 256 & 72.56 & 1.15M & 13.68B & 3.03s & 793ms\\
    3 & V2 & \checkmark & \checkmark & 256 & 72.97 & 1.02M & 12.83B & 2.98s & 786ms\\
    4 & V2 & \checkmark & \checkmark & 128 & 72.74 & 0.98M & 12.57B & 2.89s & 766ms\\
    \midrule
    5 & V3 & - & \xmark & 256 & 72.64 & 3.60M & 18.43B & 3.55s & 906ms\\
    6 & V3 & \checkmark & \xmark & 256 & 71.91 & 1.76M & 11.24B & 2.60s & 668ms\\
    7 & V3 & \checkmark & \checkmark & 256 & 72.37 & 1.63M & 10.33B & 2.55s & 659ms\\
    8 & V3 & \checkmark & \checkmark & 128 & 72.36 & 1.51M & 9.74B & 2.47s & 657ms\\
    \midrule
    \midrule
    9 & V2 0.5 & \checkmark & \checkmark & 128 & 68.57 & 0.28M & 4.00B & 1.59s & 415ms\\
    10 & V2 0.35 & \checkmark & \checkmark & 128 & 66.83 & 0.16M & 2.54B & 1.27s & 354ms\\
    \midrule
    11 & V3-Small & \checkmark & \checkmark & 128 & 68.38 & 0.47M & 2.90B & 1.21s & 327ms\\
  \bottomrule
\end{tabular}
\end{table*}
\subsection{3D Model: DeepViewAgg}
The selection of the DeepViewAgg model\cite{deepviewagg} as our 3D semantic segmentation solution is underpinned by its exceptional accuracy and efficiency, mainly when applied to the challenging KITTI-360 dataset, where it reigns as the leading benchmark performer.

One of the critical advantages of DeepViewAgg is its accessibility through a pre-trained model available on the KITTI-360 dataset's GitHub repository, which greatly facilitated our project's implementation.
DeepViewAgg has proven its versatility by successfully handling indoor and outdoor segmentation tasks. 
It has undergone rigorous testing on two pivotal datasets: the S3DIS (Stanford 3D Indoor Segmentation) and KITTI-360 datasets.

This multi-view aggregation model employs a unique approach to characterize points within point clouds based on their spatial coordinates and images based on the RGB values of their pixels while considering the intrinsic and extrinsic camera pose information. By capitalizing on the point to image pixel relationships, DeepViewAgg performs 3D point cloud semantic segmentation, leveraging features from both modalities.

The model initiates the process with a point-image mapping stage that addresses occlusions and performs efficient, GPU-accelerated point-image mapping using a streamlined Z-Buffering implementation. This mapping ensures the accurate spatial alignment of points and image pixels while handling occlusions and depth inconsistencies.

Following the point-image mapping, DeepViewAgg proceeds with the multi-view aggregation phase. This step involves attention mechanisms considering viewing conditions and relationships between multiple views. Attention mechanisms allow the model to aggregate and pool information effectively from the different camera views, improving feature representation and segmentation accuracy.

Combining point-image mapping and multi-view aggregation enables DeepViewAgg to perform 3D point cloud semantic segmentation efficiently and accurately, even when processing complex datasets at satisfactory resolutions, such as the subsampled S3DIS dataset at 5 cm resolution. 

To define the viewing conditions for each point-image pair, DeepViewAgg considers eight features, including normalized depth, local geometric descriptors [linearity, planarity, and scattering], the viewing angle to the estimated normal, the pixel row, the local density, and the occlusion rate. These viewing conditions are leveraged to construct a set of 2D feature maps through a convolutional neural network (CNN). The primary objective is to transfer features to 3D points, a process known as multi-view aggregation. An attention-based approach is employed to weigh and aggregate the features from each point \textit{p}'s viewing images.
\begin{figure}
    \centering
    \includegraphics[width=0.5\columnwidth]{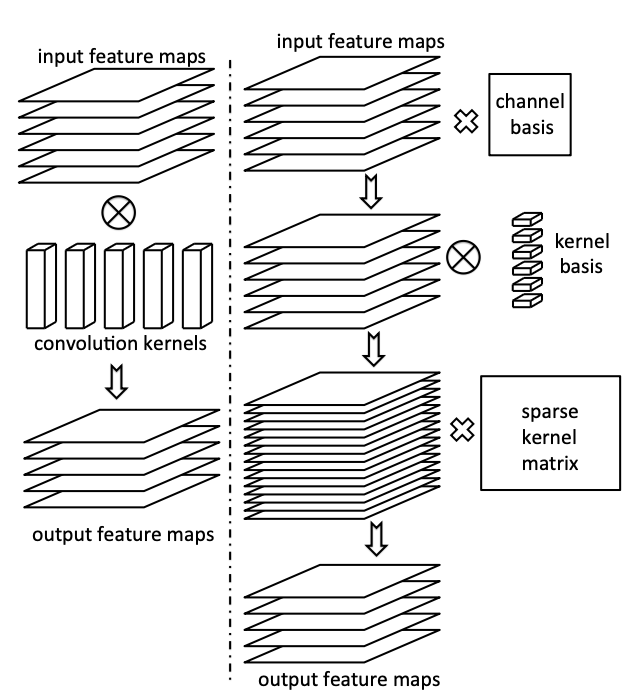}
    \caption{Sparse convolutional neural network structure compared to a regular convolution neural network. \textit{This figure is adapted with permission from the Sparse Convolutional Network paper by Liu et al. \cite{sparseconvnet}}.}
    \label{fig:sparseconvnet}
\end{figure}

DeepViewAgg relies on sparse convolutional neural networks (SCNNs), which utilize convolution operations based on sparse matrix multiplication. SCNNs differ from traditional convolutional neural network approaches as depicted in Figure \ref{fig:sparseconvnet}. As introduced by Graham et al. \cite{graham2014sparse}, sparse convolutional neural networks diverge from traditional convolutional networks in their treatment of sparse input data, where numerous values are zero. Unlike conventional CNNs operating on dense grids, sparse convolutional networks perform convolutions solely on non-zero elements, efficiently bypassing zero-valued regions. This adaptive approach significantly diminishes computational costs and memory demands, offering a more effective solution for processing sparse data in applications like point cloud data or specific types of images.

The sparse networks in DeepViewAgg employ light mapping encodings to store only compatible point-image pairs. The model's parameters are learned in two phases: initial decomposition and fine-tuning, which help create highly sparse networks, reduce computational complexity, and enhance efficiency. The sparse matrix multiplication algorithm used in sparse convolution is detailed in \cite{sparseconvnet}.

The decision to explore 3D semantic segmentation models based on sparse convolutional neural networks, such as DeepViewAgg, is rooted in their ability to manage latency, time consumption, and memory usage efficiently.
This efficiency becomes especially significant when dealing with extensive point cloud data, where the volume of points can be substantial. As demonstrated By Zhang et al.\cite{slimmer}, such networks play a pivotal role in reducing the number of points in the point cloud data without compromising accuracy, aligning with the goals of this paper.

The effectiveness of DeepViewAgg as a 3D semantic segmentation model is evident in its performance results, as reported in \cite{deepviewagg}. It establishes new state-of-the-art records in S3DIS for all six folds, achieving an mIOU of 74.7\%. Furthermore, it ranks as the second-highest performer in the fifth fold, with an mIOU of 67.2\%. DeepViewAgg outperforms other state-of-the-art 3D semantic segmentation models: PointNet++, MinkowskiNet, PointTrans., VMVF, and BPNet. In summary, \cite{deepviewagg} presents a model that significantly advances the state-of-the-art semantic segmentation for the KITTI-360 dataset, effectively integrating image and point cloud encoders.

\section{Evaluations}
\label{sec:E}
\subsection{Dataset: KITTI-360}
Our experimental investigations are conducted on the KITTI-360 dataset, chosen for its expansive collection of 3D point clouds and 2D images depicting outdoor landscapes. This unique characteristic enables the utilization of data from KITTI-360 for both 2D and 3D semantic segmentation tasks. The dataset encompasses 2D PNG images capturing raw data from a pair of 90-degree perspective stereo and 180-degree fisheye cameras.

Additionally, KITTI-360 provides 3D point cloud data obtained through advanced sensors, including the Velodyne HDL-64E and a SICK LMS 200 laser scanning unit. The dataset curates data from diverse suburban scenes in Karlsruhe, Germany, offering a rich and varied set of environmental contexts. With over 320,000 images and 100,000 laser scans covering a driving distance of 72.7 km, KITTI-360 provides a comprehensive and geolocated dataset.

The semantic label framework employed in KITTI-360 aligns with the widely adopted Cityscapes 2D dataset. Our experiments and evaluations focus on 15 classes: \textit{0} - road, \textit{1} - sidewalk, \textit{2} - building/garage, \textit{3} - wall, \textit{4} - fence/gate, \textit{5} - pole, \textit{6} - traffic light, \textit{7} - traffic sign, \textit{8} - vegetation, \textit{9} - terrain, \textit{10} - person, \textit{11} - car, \textit{12} - truck, \textit{13} - motorcycle, \textit{14} - bicycle. Notably, the geolocalization of all frames in the dataset is accurately established with the assistance of OpenStreetMap \cite{kitti360}.

\subsection{Metrics}
We assess the semantic segmentation accuracy using Intersection over Union (IoU), defined in Equation \ref{eq:IoU}

\begin{equation}
IoU = \frac{TP}{TP + FP + FN}
\label{eq:IoU}
\end{equation}

Here, TP represents True Positives, FP stands for False Positives, and FN corresponds to False Negatives. IoU provides insights into the degree of overlap between the predicted and ground truth pixel labels.  

To assess the computational efficiency of the semantic segmentation process, we utilize a single time metric and three key memory metrics: Run time, Program memory, Model memory, and Runtime memory. The memory profiling is performed using the $psutil$ library.

\begin{itemize}
    \item \textbf{Run Time}: The average run time measurements represent the duration taken to calculate the output from applying the model on the input image, which is averaged over the evaluation runs.
    \item \textbf{Program Memory}: The program memory characterizes the executed program's memory. For DeepViewAgg, this pertains to the 3D inference script employed to evaluate the model across the validation and test datasets. In our approach, program memory is further categorized into the inference script for 2D semantic segmentation evaluation, the script executing the extrusion on 2D segmented data, and the inference script used to assess the 3D model on truncated data for both datasets.
    \item \textbf{Model Memory}: Model memory quantifies the memory utilized by loading the model into the program. For DeepViewAgg, this is associated with the 3D model, serving as our baseline. In contrast, our approach involves the consideration of both the 2D and 3D models used.
    \item \textbf{Runtime Memory}: The average runtime memory denotes the memory required to execute the model on input data and generate an output. This measurement aligns with the time measurement previously discussed.
\end{itemize}

All averages are calculated over 20 runs of running the model over the entire validation and test datasets, each, for 2D semantic segmentation and over five runs over the entire validation and test datasets for 3D semantic segmentation.
These memory and time metrics collectively contribute to a comprehensive assessment of the model's accuracy and efficient resource utilization for semantic segmentation tasks.

\subsection{Baseline}

The current state-of-the-art approach for identifying objects associated with specific class labels within a point cloud entails performing 3D semantic segmentation on the entire point cloud. Since this is a widely adopted method, we utilize it as our baseline for the comparative assessment of our novel method. We aim to establish a methodology for resource and time-efficient 3D semantic segmentation with minimal to no loss in validation accuracy.

To establish a performance baseline for our approach, we employ DeepViewAgg on the complete point cloud. This baseline serves as a reference for comparing our proposed method. DeepViewAgg is applied to both the validation dataset, where we compute accuracy metrics, primarily Intersection over Union (IOU) per class, and the validation and test datasets for evaluating time and memory consumption, as discussed in the metrics subsection above.

To advance the efficiency and practicality of 3D semantic segmentation, it is important to benchmark against the best-performing models in the field. 
DeepViewAgg serves as the current state-of-the-art approach for 3D semantic segmentation on the KITTI-360 dataset and hence represents the pinnacle of accuracy and performance for segmentation in this area. 
The selection of DeepViewAgg as the baseline model to compare our approach's performance against was strategically chosen for the following considerations:

\begin{itemize}
    \item \textbf{Benchmarking Against the Best}: DeepViewAgg's status as the leading model for 3D semantic segmentation on the KITTI-360 dataset provides a stringent benchmark. We are able to assess our approach's ability to challenge and potentially surpass the current best in terms of efficiency with minimal compromise in accuracy by comparing our approach against DeepViewAgg.
    \item \textbf{Focus on Efficiency and Memory Consumption}: While DeepViewAgg excels in achieving the current best IoU accuracy, it is also accompanied by computational and memory requirements. Our work is motivated by the goal to substantially improve upon these aspects—reducing time and memory consumption—while striving to maintain, and in some cases improve, segmentation accuracy. By setting DeepViewAgg as our benchmark, we aim to demonstrate tangible advancements in these critical areas.
    \item \textbf{Relevance and Comparability}: The choice of DeepViewAgg facilitates a focused and relevant comparison. Given its established excellence in IoU accuracy on the KITTI-360 dataset, any observed improvements in efficiency and memory usage by our approach can be directly attributed to the technical contributions of our methodology. This not only emphasizes the practical contributions of our work but also enhances the comparability of our results in the broader research community.
    \item \textbf{Establishing a Proportional Baseline}: By benchmarking against the state-of-the-art, we set a proportional baseline that allows for extrapolating the efficiency gains of our approach to other segmentation models for the KITTI-360 dataset. Improved efficiency by our method over DeepViewAgg suggests that similar benefits could be achieved when applied to other contexts, thereby broadening the impact and applicability of our findings.
    \item \textbf{Facilitating Clear Insights}: Direct comparison with DeepViewAgg ensures a straightforward evaluation of our approach's performance. It provides clear insights into how the proposed methodology fares directly against the state-of-the-art, offering a concise understanding of its potential to redefine efficiency standards in 3D semantic segmentation.
\end{itemize}

In essence, using DeepViewAgg as the sole baseline for comparison is a strategic decision, designed to highlight the specific advancements our approach offers in efficiency and memory usage while ensuring that these improvements are measured against the highest standard of accuracy currently available. 
This approach allows us to present a clear, focused, and compelling case for our approach's advantages within the evolving and growing state of 3D semantic segmentation research.

\subsection{Experimental Testbed}
Our experiments were executed on a Linux-based system featuring dual Intel Xeon Silver 4214R CPUs with 48 cores and 250GB of RAM. Three NVIDIA RTX A6000 GPUs with 49GB of combined memory were employed. The system ran Ubuntu 20.04.6 LTS, and Python 3.7.9 was used for model development.

\subsection{Hyperparameters}
MobileNetv3 was employed for semantic segmentation on the KITTI-360 dataset with specific hyperparameters. 
A batch size of 8 was used for training, and the initial learning rate was set to 0.01. 
The training process encompassed 240 epochs, and optimization was performed using the Adam optimizer. 
Data augmentation techniques, like random horizontal flipping and color jittering, were applied to enhance the model's robustness and generalization. 

DeepViewAgg was trained on the KITTI-360 dataset with a batch size 16 and an initial learning rate of 0.001. 
The training spanned 80 epochs, and the Adam optimizer was employed for optimization. 
Data preprocessing included point cloud subsampling at 5 cm intervals with cylindrical sampling. 
During testing, the model's performance was assessed using standard semantic segmentation metrics such as mIOU and accuracy; however, the IOU per class is specifically reported in this paper.

These carefully tuned hyperparameters played a pivotal role in achieving the optimal performance of MobileNetv3 and DeepViewAgg on the challenging KITTI-360 dataset, ensuring their efficiency and accuracy in 3D semantic segmentation.

\section{Results and Discussion}
\label{sec:RD}

\begin{table*}
\centering
\caption{Semantic segmentation accuracy (\textit{IoU} measured in \%) results on the validation dataset compared between DeepViewAgg (\textit{DVA}) and our approach (\textit{OA}). Class label definitions are as follows: \textit{0} - road, \textit{1} - sidewalk, \textit{2} - building/garage, \textit{3} - wall, \textit{4} - fence/gate, \textit{5} - pole, \textit{6} - traffic light, \textit{7} - traffic sign, \textit{8} - vegetation, \textit{9} - terrain, \textit{10} - person, \textit{11} - car, \textit{12} - truck, \textit{13} - motorcycle, and \textit{14} - bicycle.}
\label{tab:accres}
\resizebox{\textwidth}{!}{%
  \begin{tabular}{cc|c|c|c|c|c|c|c|c|c|c|c|c|c|c|c|}
\cline{3-17}
    & & \multicolumn{15}{c|}{\textbf{IoU per class label (\%)}} 
    \\ \cline{3-17} 
    & & 0 & 1 & 2 & 3 & 4 & 5 & 6 & 7 & 8 & 9 & 10 & 11 & 12 & 13 & 14  \\ \hline
\multicolumn{1}{|c|}{\multirow{2}{*}{\textbf{Model}}} & DVA & \textbf{93.22} & \textbf{76.39} & 85.89 & \textbf{49.52} & \textbf{44.61} & \textbf{57.60} & 0.00 & \textbf{47.72} & \textbf{87.90} & \textbf{69.61} & 58.29 & 94.01 & 56.88 & \textbf{55.03} & 24.51 
    \\ \cline{2-17} 
\multicolumn{1}{|c|}{} & OA & 93.11 & 76.21 & \textbf{85.91} & 49.43 & 43.98 & 57.35 & \textbf{8.78} & 47.53 & 87.88 & 69.43 & \textbf{62.92} & \textbf{94.03} & \textbf{65.36} & 54.75 & \textbf{26.14} \\ \hline
\end{tabular}%
}
\end{table*}

\begin{table*}
  \centering
  \caption{Time latency and memory consumption results are tabulated for semantic segmentation on the validation and test datasets compared between DeepViewAgg and our approach. The abbreviations used in the table are as follows: Average Run time (\textit{Run time}), Program memory (\textit{Program}), Model memory (\textit{Model}), Average Runtime memory (\textit{Runtime Mem.}), Validation (\textit{Val.}), and 2D-3D Extrusion Step (\textit{2D-3D Ext.}).}
  \label{tab:time_mem}
  \resizebox{\textwidth}{!}{%
  \begin{tabular}{cccc|c|c|c|c|}
    \cline{5-8}
    & & & & \textbf{Run time (ms)} & \textbf{Program (MB)} & \textbf{Model (MB)} & \textbf{Runtime Mem. (MB)} \\
    \hline
    \multicolumn{1}{|c|}{\multirow{5}{*}{\textbf{Model}}} & DeepViewAgg & \multicolumn{2}{|c|}{Val.} & $998.391 \pm 28.602$ & $\bm{552.100 \pm 2.177}$ & $660.876 \pm 217.794$ & $157.472 \pm 3.819$ \\
    \cline{2-8}
    \multicolumn{1}{|c|}{} & \multicolumn{1}{c|}{\multirow{4}{6em}{Our Approach}} & \multicolumn{1}{c|}{\multirow{4}{1.5em}{Val.}} & \multicolumn{1}{c|}{2D} & $138.661 \pm 24.367$ & $215.928 \pm 0.998$ & $17.732 \pm 0.113$ & $0.470 \pm 41.830 \times 10^{-3}$ \\
    \cline{4-8}
    \multicolumn{1}{|c|}{} & \multicolumn{1}{c|}{} & \multicolumn{1}{c|}{} & \multicolumn{1}{c|}{2D-3D Ext.} & $183.849 \pm 49.719$ & $317.686 \pm 0.531$ & - & $2.369 \times 10^{-3} \pm 1.346$ \\
    \cline{4-8}
    \multicolumn{1}{|c|}{} & \multicolumn{1}{c|}{} & \multicolumn{1}{c|}{} & \multicolumn{1}{c|}{3D} & $452.156 \pm 176.071$ & $315.647 \pm 83.504$ & $383.600 \pm 229.900$ & $74.053 \pm 26.701$ \\
    \cline{4-8}
    \multicolumn{1}{|c|}{} & \multicolumn{1}{c|}{} & \multicolumn{1}{c|}{} & \multicolumn{1}{c|}{Total} & $\bm{774.666 \pm 250.157}$ & $849.261 \pm 85.033$ & $\bm{401.332 \pm 230.013}$ & $\bm{74.525 \pm 28.089}$ \\
    \hline 
    \multicolumn{8}{|c|}{}\\[-0.75em]
    \hline
    \multicolumn{1}{|c|}{\multirow{5}{*}{\textbf{Model}}} & \multicolumn{1}{c|}{DeepViewAgg} & \multicolumn{2}{c|}{Test} & $933.272 \pm 157.542$ & $\bm{558.608 \pm 2.208}$ & $591.794 \pm 213.986$ & $115.019 \pm 0.754$ \\
    \cline{2-8}
    \multicolumn{1}{|c|}{} & \multicolumn{1}{c|}{\multirow{4}{6em}{Our Approach}} & \multicolumn{1}{c|}{\multirow{4}{1.5em}{Test}} & 2D & $171.071 \pm 25.781$ & $216.844 \pm 0.291$ & $17.816 \pm  0.101 $ & $1.358 \pm  0.059$ \\
    \cline{4-8}
    \multicolumn{1}{|c|}{} & \multicolumn{1}{c|}{} & \multicolumn{1}{c|}{} & 2D-3D Ext. & $178.975 \pm 49.461$ & $318.690 \pm 0.676$ & - & $12.093 \times 10^{-3} \pm 2.945$ \\
    \cline{4-8}
    \multicolumn{1}{|c|}{} & \multicolumn{1}{c|}{} & \multicolumn{1}{c|}{} & 3D & $342.785 \pm 218.690$ & $320.319 \pm 98.016$ & $289.854 \pm 86.473$ & $63.475 \pm 14.847$ \\
    \cline{4-8}
    \multicolumn{1}{|c|}{} & \multicolumn{1}{c|}{} & \multicolumn{1}{c|}{} & Total & $\bm{692.831 \pm 293.392}$ & $855.853 \pm 98.983$ & $\bm{307.670 \pm  86.574} $ & $\bm{64.845 \pm  17.851}$ \\
    \cline{2-8}
    \hline
\end{tabular}
}
\end{table*}
We compare our approach against the baseline discussed in Section \hyperref[sec:E]{4}, DeepViewAgg, by evaluating Intersection over Union (IoU) accuracy results per class. Table~\ref{tab:accres} presents the results for the 15 classes mentioned. Our approach demonstrates improved accuracy for 6 out of 15 class labels [2 (building/garage), 6 (traffic light), 10 (person), 11 (car), 12 (truck), and 14 (bicycle)] while maintaining accuracy within 1\% of DeepViewAgg for the remaining class labels.

Existing work such as that of DASS by Unal et al. \cite{unal2021DASS} and surveys on point cloud segmentation like those by Zhang et al. \cite{zhang2023survey} and Xie et al. \cite{xie2020review} provide evidence that the density of points as they relate to class labels in a point cloud can affect the accuracy of performing segmentation for those class labels.
Specifically, a reduction in the density of points associated with a class label can negatively affect the segmentation accuracy of the model for the class labels in question.
Our approach mitigates this challenge by utilizing 2D semantic segmentation and our extrusion process to focus the segmentation classification process on regions of the point cloud that are associated with the class labels in question.
By performing segmentation on a subspace of the points in the point cloud, we are able to increase the density of points associated with a selected class label in the region of points where we perform 3D semantic segmentation.
We hypothesize that this allows us to improve the segmentation accuracy in the cases where we consider class labels with lower densities in the complete point cloud, namely class labels 6 and 12 which have the highest accuracy discrepancies as compared to DeepViewAgg, shown in Table \ref{tab:accres}.

In addition, our approach is dependent on a sparse convolutional neural network (SCNN) architecture in performing 3D semantic segmentation. 
In our SCNN architecture, sparsification is applied to the data points that are highlighted by our 2D semantic segmentation and extrusion processes. 
This ensures that points belonging to the target class labels are prioritized and less likely to be overlooked. 
This is particularly beneficial for class labels that are sparsely represented in the original point cloud.
We hypothesize that this method enhances efficiency and accuracy, particularly for class labels that are less represented in the original point cloud.
We hypothesize that the number of point clouds in our training dataset that contain points associated with a given class label and the respective point density for that class label in the point clouds affect the accuracy of performing segmentation for that class label and especially affect the discrepancies in accuracy between our approach and the baseline, DeepViewAgg, as can be seen in Table \ref{tab:accres}.
By adapting our segmentation strategies to account for variations in point density, our approach showcases its ability to better capture the details necessary for accurate class label identification, even when the number of points is limited.
As a part of our future work, we intend to look deeper into the relationships among point density, number of point clouds with points associated with target class labels, and the accuracy we see from performing segmentation with a model based on an SCNN architecture, as this has the power of enabling improved understanding of the reasoning behind the accuracy values we see in Table \ref{tab:accres}.

Table~\ref{tab:time_mem} provides insights into time and memory consumption, averaged over five runs for validation and test datasets. Our approach exhibits reduced processing time ($774.666 \pm 250.157$ ms for validation, $692.831 \pm 293.392$ ms for test) compared to DeepViewAgg ($998.391 \pm 28.602$ ms for validation, $933.272 \pm 157.542$ for test), contributing to a 1.289x speedup for validation and a 1.347x speedup for test. Regarding memory consumption, our approach utilizes less overall model memory ($401.332 \pm 230.013$ MB for validation, $307.670 \pm 86.574$ MB for test) than DeepViewAgg ($660.876 \pm 217.794$ MB for validation, $591.794 \pm 213.986$ MB for test), which amounts to 39.273\% decrease in memory usage for the validation dataset and 48.011\% decrease for the test dataset. In addition, our approach uses less runtime memory ($74.525 \pm 28.089$ MB for validation, $64.845 \pm  17.851$ MB for test) than DeepViewAgg ($157.472 \pm 3.819$ MB for validation, $115.019 \pm 0.754$ MB for test), a 52.674\% decrease for validation and 43.622\% decrease for test.

This large reduction in memory consumption and inference run time latency for our approach as compared to DeepViewAgg serves as a proof-of-concept that our approach yields benefits for potential deployment on resource-constrained devices and in real-time applications, where latency in inference time and consumption of memory are important to consider when performing 3D semantic segmentation.
In addition, our approach showcases improved accuracy for the traffic light, person, car, truck, and bicycle classes. 
These classes are crucial for segmentation in dynamic environments, where objects within these categories are in motion, as seen in real-time applications like autonomous driving.
For these applications, utilizing our 3D semantic segmentation approach can prove highly useful as our approach can perform accurate segmentation over the existing state-of-the-art for these classes while ensuring high performance in run time latency and memory consumption.

One limitation of our approach is that there is an increase in program memory usage, which can be attributed to incorporating 2D semantic segmentation and our extrusion algorithm. 
However, the overall increase remains manageable in comparison to the speedup and reduction in other memory usage attained by our approach, amounting to 53.824\% increase in program memory consumption for validation and 53.212\% increase for test.
Future work will aim to address this limitation by integrating quantization and other optimizations to the programs to reduce the data associated with storage and memory allocation during consecutive runs through the model.

As we compare our approach to DeepViewAgg, there are multiple noteworthy considerations of our approach as it addresses the limitations of the DeepViewAgg model discussed in Section \hyperref[sec:RWM]{2}.
In the context of computational efficiency and memory use, our approach optimizes the integration of 2D semantic segmentation with 3D point clouds which helps to reduce the inference run time overhead and memory usage by streamlining the process of extrapolating 2D semantic segmentation results to 3D spaces.
DeepViewAgg is dependent on the availability and quality of multiple 2D views for performing its aggregation, however our method improves robustness of 3D semantic segmentation by enhancing efficiency and accuracy even with limited or suboptimal 2D views, with the only consideration being the ability to perform 2D semantic segmentation on the view to extrude the output to the 3D point cloud.
By focusing on optimizing the extrusion process and the integration of 2D semantic segmentation into 3D point clouds, our approach also provides the potential for improved generalization across diverse scenarios, particularly by addressing the limitations in capturing complex 3D structures from 2D views.
Our approach also efficiently leverages 2D semantic segmentation results and hence minimizes the need for extensive data aggregation, which can help to reduce latency, making it more applicable for real-time or near-real-time applications.

\section{Conclusion and Future Work}
\label{sec:CFW}
In this research paper, we propose and evaluate a novel approach to enhance efficiency for 3D semantic segmentation, explicitly targeting the KITTI-360 dataset. Our method employs a hybrid vision technique to streamline computational demands, resulting in notable reductions in processing time and memory consumption compared to the DeepViewAgg  when applied to the complete point clouds of the KITTI-360 dataset.

Our process begins with 2D semantic segmentation on RGB-encoded images correlated with the 3D point clouds. The resultant 2D segmented data allows us to identify essential image pixels associated to target class labels. We utilize the essential image pixels to narrow the point cloud subspace to points associated with specific target class labels. This narrowing process significantly reduces processing time and memory usage during 3D semantic segmentation. Our approach ensures that these efficiency improvements maintain the accuracy of our results. In comparative evaluations against the baseline 3D semantic segmentation model, our method outperforms the accuracy of the baseline for \textbf{6} out of the \textbf{15} classes for the validation dataset while establishing a \textbf{1.289x} speedup and about a \textbf{52.674\%} reduction in memory usage for the validation dataset and a \textbf{1.347x} speedup and at least a \textbf{43\%} reduction in memory usage for the test dataset.

While our research has made notable strides in enhancing the efficiency of 3D semantic segmentation, there are several promising avenues for future exploration and refinement:
\begin{itemize}
    \item Optimization of the hybrid vision technique, focusing on minimizing information loss while maximizing computational gains during the 2D-3D transition.
    \item Extension of the approach to diverse datasets and domains to validate its robustness and versatility.
    \item Exploration applicability of our approach in dynamic environments with moving objects.
    \item Enhancing the model's adaptability to different hardware platforms can ensure the reproducibility of observed speedup and memory improvements across various devices and systems with limited resources.
    \item Integration of  real-time feedback mechanisms and adaptability features for user-defined class labels holds promise for customizing the approach to specific use cases.
\end{itemize}

Our work on efficient 3D semantic segmentation through hybrid vision techniques holds significant implications for various practical applications. Highly accurate point cloud data is valuable, especially for performing 3D semantic segmentation, due to its efficacy in improving accuracy due to the quantity and quality of the data provided to such segmentation models. This methodology can be extended to autonomous vehicles and robotics, where real-time segmentation is crucial for safe navigation and enhanced user experiences. Moreover, our approach can facilitate precise tumor segmentation in medical imaging, where improved and real-time results aid diagnosis and treatment planning, especially in surgery and therapeutic approaches. The versatility of our method also makes it applicable in areas like environmental monitoring, industrial automation, and cultural heritage preservation, where 3D semantic segmentation is valuable for data analysis and decision-making.



\end{document}